\documentclass[runningheads]{llncs}
\usepackage{graphicx}
\usepackage{tikz}
\usepackage{comment}
\usepackage{amsmath,amssymb} % define this before the line numbering.
\usepackage{color}

\usepackage{booktabs}
\usepackage{floatrow}
\usepackage{multicol}
\usepackage{tabularx}
\usepackage{adjustbox}
\usepackage{multirow}
\usepackage{makecell}
\usepackage{xcolor}
\usepackage{placeins}
\usepackage{caption}
\usepackage{subcaption}

\graphicspath{ {./images/} }

\usepackage[accsupp]{axessibility}  % Improves PDF readability for those with disabilities.

\usepackage[capitalize]{cleveref}

\newfloatcommand{capbtabbox}{table}[][\FBwidth]

\crefname{section}{Sec.}{Secs.}
\Crefname{section}{Section}{Sections}
\Crefname{table}{Table}{Tables}
\crefname{table}{Tab.}{Tabs.}

\newcommand{\nj}[1]{\textcolor{black}{#1}}

\begin{document}
% \linenumbers
\pagestyle{headings}
\mainmatter
\def\ECCVSubNumber{5}  % Insert your submission number here

\title{Conservative Generator, Progressive Discriminator: Coordination of Adversaries in Few-shot Incremental Image Synthesis} % Replace with your title
% \title{Smoothing the Generative Latent Space with Mixup-based Distance Learning} % Replace with your title

\begin{comment}
\titlerunning{ECCV-22 submission ID \ECCVSubNumber} 
\authorrunning{ECCV-22 submission ID \ECCVSubNumber} 
\author{Anonymous ECCV submission}
\institute{Paper ID \ECCVSubNumber}
\end{comment}
%******************

% \begin{comment}
\titlerunning{Conservative Generator, Progressive Discriminator}
\author{Chaerin Kong \and
Nojun Kwak}
\authorrunning{C. Kong et al.}
\institute{Seoul National University \\
\email{\{veztylord,nojunk\}@snu.ac.kr}}
% \end{comment}
%******************
\maketitle

\begin{abstract}
  %The capacity to learn incrementally from an online stream of data is an envied trait of human learners, as deep neural networks typically suffer from catastrophic forgetting and stability-plasticity dilemma. 
  %Several works have previously explored incremental few-shot learning, a task with greater challenges due to data constraint, mostly in classification setting with mild success. 
  In this work, we study the underrepresented task of generative incremental few-shot learning. To effectively handle the inherent challenges of incremental learning and few-shot learning, we propose a novel framework named ConPro that leverages the two-player nature of GANs. Specifically, we design a conservative generator that preserves past knowledge in a parameter- and compute-efficient
%   \nj{in an} : parameter-efficient, compute-efficient 를 의도했는데, 표현 조금 생각해보겠습니다~
  %efficient 
  manner, and a progressive discriminator that learns to reason semantic distances between past and present task samples, minimizing overfitting with few data points and pursuing good forward transfer. 
  We present experiments to validate the effectiveness of the proposed framework.
%   \nj{Experimental results} clearly demonstrate the effectiveness of our approach.

%\dots
\keywords{Generative Adversarial Networks (GANs), Few-shot Image Synthesis, Incremental Learning}
\end{abstract}

\section{Introduction}
\label{sec:intro}

The ability to learn from a continuum of data is essential not only for the academic pursuit of human-like AI but also for many practical scenarios where the training dataset is imperfect and prone to distribution shifts. Though incremental learning (IL) has long been an important research topic, catastrophic forgetting~\cite{mccloskey1989catastrophic} still poses grave challenges that even the state-of-the-art methods fail to %address 
overcome perfectly. Moreover, the concept of incremental learning has mainly been studied in the classification context~\cite{kirkpatrick2017overcoming,li2017learning}, leaving other downstream tasks such as generative modeling relatively less explored.

Meanwhile, as deep neural networks are notoriously data hungry, data-efficient training has gained increasing attention from the community in an attempt to broaden their applications and imitate human intelligence~\cite{brown2020language,jang2023unifying,jang2023self,kong2023analyzing}. Data-efficient training, also known as low-shot or few-shot learning, has been actively studied in a wide variety of task settings including classification~\cite{finn2017model,snell2017prototypical}, object detection~\cite{kang2019few}, 3D view synthesis~\cite{yu2021pixelnerf,song2022towards} and content generation~\cite{liu2020towards,li2020few,ojha2021few,tseng2021regularizing,kong2021smoothing,kong2022few,kong2023leveraging,lee2023aadiff}. Few-shot image synthesis, in specific, aims to generate diverse samples of reasonable quality by suppressing memorization. Several recent works~\cite{kong2022few,liu2020towards,cui2021genco} demonstrated promising results with as few as 10 to 100 training samples.

% typically relies on transfer learning due to its innate difficulty, but recent works~\cite{MDL, GenCo, FastGAN} showed promising results even without pretraining.

In this work, we attempt to bridge the gap between the two core components of human intelligence, namely incremental learning and few-shot learning, and explore incremental few-shot learning of generative models.
This task setting imposes two fundamental challenges: catastrophic forgetting and overfitting. 
In order to overcome this pair of obstacles, we leverage the two-player nature of GANs in a \textit{divide-and-conquer} manner. In our framework, the conservative generator adopts task-specific modular expansion with minimal memory- and compute-growth to preserve past knowledge and serve as a memory replay generator. The progressive discriminator, on the other hand, minimizes overfitting by actively engaging in in-domain (current task) and cross-domain (past tasks) distance learning, where previous task samples are generated by the conservative generator.
% motivated by techniques introduced in recent self-supervised learning literature~\cite{SimCLR}. 
Combining the pair, our method (ConPro) successfully handles the two challenging problems at once thanks to the reinforcing synergy of the adversaries. 

% Our key insight is that in incremental few-shot learning (IFL), only the generator is responsible for past tasks and hence the role of discriminator has been largely overlooked~\cite{GANmemory, CAMGAN}. 

Our key insight is that conventional IL methods largely overlook the role of discriminator since the generator is the one responsible for past tasks. Nevertheless, as incremental few-shot learning (IFL) poses additional data constraint, we believe that the discriminator \textit{matters} and thus should be actively engaged. Empirical advantages of ConPro further supports our claim.

% The primary goal of our work is to demonstrate that the discriminator \textit{matters} in IFL, and thus \textit{can and should be} actively engaged to boost sample quality and diversity. Empirical advantages of ConPro further supports our claim.

% Motivated by recent self-supervised learning literature~\cite{SimCLR, MoCo}, we apply distance learning regularizations that employ both current and past task samples to mitigate memorization and strengthen semantic reasoning under data constraint. 

% has greatly limited its application, and even the state-of-the-art methods are still far from conquering it. 

% \section{Related Works}
% \label{sec:rel}

\section{Approach}
\label{sec:app}

We consider the class-incremental setting where our conditional generative model learns to generate samples from an incoming sequence of classes while preserving the knowledge about past classes. One benefit of this problem setting is that the generative model can be further employed as a generative replay for continual learning of other downstream tasks~\cite{shin2017continual}. In \cref{subsec:gen}, we discuss the parameter- and compute-efficient design of our conservative generator that employs factorized modulation~\cite{skorokhodov2021adversarial}. \cref{subsec:disc} describes the discriminator, which actively compares and contrasts current class samples and past samples to learn stronger visual reasoning as the training proceeds (hence progressive), and provide a better guidance to the generator. Based on \cref{subsec:disc}, we introduce a simple yet effective weight initialization technique based on discriminator's domain distance evaluation that fosters performance and convergence.

\subsection{Adaptive Factorized Modulation}
\label{subsec:gen}

We propose a task-specific weight modulation technique inspired by Factorized Multiplicative Module~\cite{skorokhodov2021adversarial}. Following the convention of IFL that typically assumes a large base task and the ensuing few-shot tasks, we fix the parameters learned from the base task and only train the task-specific modulation parameters for each task. 
% As these task-specific parameters are used to modulate the base parameters, they can be low-rank factorized with little loss~\cite{skorokhodov2021adversarial} to save memory. 

\begin{figure}[t]
  \includegraphics[width=\linewidth]{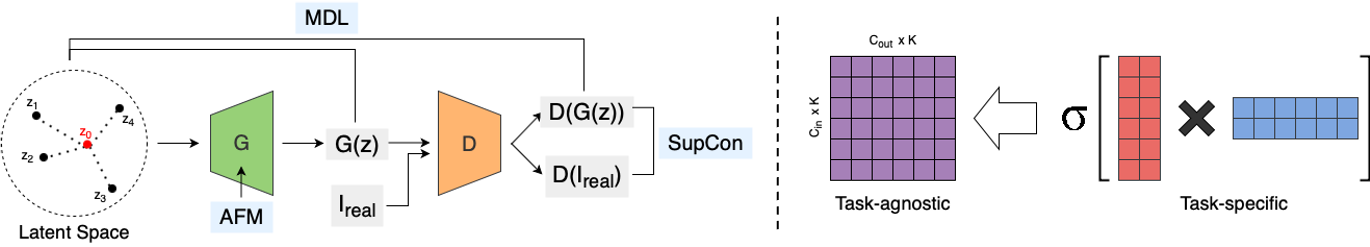}
%   \vspace{-20pt}
  \centering
%   \captionsetup{width=0.8\linewidth}
  \caption{\small
  ConPro overview (left) and Adaptive Factorized Modulation (\nj{AFM}, right).
  }
  \vspace{-3mm}
\label{fig:main}
\end{figure}

Formally, let $F \in \mathbb{R}^{{c_{out}}\times{c_{in}}\times k \times k}$ be a convolutional layer parameters with kernel size $k$ learned from the base task. We reshape as ($c_{out}\times k, c_{in}\times k$) and apply task-specific modulation as element-wise multiplication. Our modulation parameters are low-rank factorized with two matrices of shape ($c_{out} \times k$, $r$), ($r$, $c_{in} \times k$) where $r$ refers to the rank, and an ensuing sigmoid activation ensures their range. Graphical illustration is provided in \cref{fig:main}.
% As we later show in \cref{sec:exp}, our approach outperforms previous model expansion baselines with better memory and compute efficiency.
% Compared to GAN-memory~\cite{gan-memory}, ours is parameter-efficient thanks to low-rank factorization. While CAM-GAN~\cite{cam-gan} shows reasonable memory-efficiency, the computation burden is significant as they add task-specific convolution layers. As we show in \cref{sec:exp}, ours 

% \begin{figure}[t]
% \centering
% \begin{subfigure}{.6\textwidth}
%   \centering
%   \includegraphics[width=.95\linewidth]{figure/Conpro_Dist.drawio.png}
%   \caption{Overview of ConPro.} 
%   \label{fig:main1}
% \end{subfigure}%
% \begin{subfigure}{.4\textwidth}
%   \centering
%   \includegraphics[width=.95\linewidth]{figure/Conpro_AFM.drawio.png}
%   \hspace{10pt}
%   \caption{Adaptive Factorized Modulation (AFM).}
%   \label{fig:main2}
% \end{subfigure}
% \caption{Our method outline.}
% \label{fig:main}
% \end{figure}

\vspace{-3mm}
\subsection{Discriminator Distance Learning}
\label{subsec:disc}

With the help of knowledge preserving generator, our discriminator can be trained with both present and past task samples. As opposed to previous works~\cite{cong2020gan,varshney2021cam} that left it untouched, we claim that the discriminator plays an important role under additional dataset constraint and hence deserves modifications. Inspired by self-supervised learning~\cite{chen2020simple,he2020momentum} and few-shot image generation~\cite{kong2021smoothing}, we introduce two distance learning objectives for in-domain and cross-domain:

\noindent \textbf{In-domain distance learning} adopts the formulation of MDL~\cite{kong2021smoothing} to smooth the latent space. We sample latent mixup coefficients from \textit{e.g.,} Dirichlet, generate latent-mixup samples and enforce them to have semantic similarities proportional to the mixup ratio. Formally, the objective is as follows:

\vspace{-15pt}

\begin{align}
    \mathcal{L}_{MDL}^G &= \mathbb{E}_{z \sim p_z(z), \mathbf{c} \sim Dir(\mathbf{1})}[KL(P^l||Q)], \\
    \label{eqn:ours_source} 
    P^l &= \text{softmax}(\{\text{sim}(G^l(z_0), G^l(z_i))\}_{i=1}^N), \\
    Q &= \text{softmax}(\{c_i\}_{i=1}^N)
    % z_0 &= \sum_{i=1}^N c_i z_i, \quad \mathbf{c} \sim Dir(\alpha_1, \cdots , \alpha_N).
\end{align}

\vspace{-5pt}
\noindent
where $l$ indicates activation layer index, $Dir(\mathbf{1})$ refers to the Dirichlet distribution with all-1 parameters and $\mathbf{c} \triangleq [c_1, \cdots, c_N]^T$. 
% We train our generator \nj{by the combination of} adversarial loss and MDL loss. 
Similar regularization is imposed on the discriminator as well with an additional linear projection layer denoted \textit{proj} as in \cite{chen2020simple}.

\vspace{-18pt}

\begin{align}
    \mathcal{L}_{MDL}^D &= \mathbb{E}_{z \sim p_z(z), \mathbf{c} \sim Dir(\mathbf{1})} [KL(R||Q)], \\
    R &= \text{softmax}(\{\text{sim}(proj(d_0), proj(d_i))\}_{i=1}^N).
\end{align}

%\nj{prod 가 뭔지 설명이 안 되어 있음. Fig. 1의 right side에 대한 설명 필요. }

\vspace{-5pt}

\noindent \textbf{Cross-domain distance learning} imposes supervised contrastive learning objective~\cite{khosla2020supervised} between current real samples and past genenerated images where all samples belonging to the same class are regarded as \textit{positive samples}. As naive data augmentations can hurt synthesis quality, we simply omit them and leave the incorporation for future work. Our final objective is as follows:

\vspace{-10pt}
\begin{align}
    \mathcal{L}^G &= \mathcal{L}^G_{adv} + \lambda^G_{MDL} \mathcal{L}^G_{MDL}, \\
    \mathcal{L}^D &= \mathcal{L}^D_{adv} + \lambda^D_{MDL} \mathcal{L}^D_{MDL}  + \lambda_{SupCon} \mathcal{L}_{SupCon}.
\end{align}

\vspace{-3mm}
\subsection{Distance Aware Initialization (DAI)}
\label{subsec:dai}

Good parameter initializations can improve model convergence and the overall performance greatly~\cite{finn2017model,varshney2021cam}. Since our discriminator is trained to reason about semantic distances between cross-domain samples, 
we leverage this capacity to initialize generator modulation parameters with those from the most relevant past task.
% we have our discriminator evaluate distances for each past task with respect to current task, and initialize the generator modulation parameters with the most relevant task parameters.
% we have our generator modulation parameters initialized as those from the most relevant task, identified by the discriminator's domain distance evaluation. 
That is, we draw a batch of current real and past generated samples, compute their distances in the discriminator feature space and choose the task with minimal feature distance to the current batch of real samples. As we show in \cref{sec:exp}, DAI stabilizes training and improves the final performance. Also we note that it is much simpler and faster than TSL in \cite{varshney2021cam} that computes Fisher Information Matrix of generator parameters to find the relevant task.

\section{Experiments}
\label{sec:exp}

We conduct experiments using Animal-Face dataset that consists of roughly 100 samples per class. We simulate the incremental setting with 7 randomly sampled classes, and use Celeb-A pretrained GP-GAN as the backbone following \cite{varshney2021cam,cong2020gan}. 
% We re-implement CAM-GAN~\cite{camgan} and report Frechet Inception Distance (FID)~\cite{fid} for fidelity and diversity metric.

\begin{table}[t]
% \vspace{-12pt}
\caption{Quantitative results on few-shot incremental generation with randomly sampled Animal Face Dataset. Ours provides consistent gains over the baseline.}
\centering
\resizebox{0.7\linewidth}{!}{\scriptsize
\begin{tabular}{l|lllllll}
\Xhline{3\arrayrulewidth}
\multicolumn{1}{c|}{FID} & \multicolumn{1}{c}{Task 1} & \multicolumn{1}{c}{Task 2} & \multicolumn{1}{c}{Task 3} & \multicolumn{1}{c}{Task 4} & \multicolumn{1}{c}{Task 5} & \multicolumn{1}{c}{Task 6} & \multicolumn{1}{c}{Task 7} \\ 
\hline
CAM-GAN~\cite{varshney2021cam}             & 254.5                      & 213.3                      & 257.6                      & 303.6                      & 199.3                      & 253.2                      & 235.2                      \\ \hline
+ AFM             & 232.8                      & 229.2                      & 278.9                      & 175.4                      & 292.6                      & 241.6                      & 164.8                      \\
+ MDL                    & 184.0                      & 173.7                      & \textbf{229.5}                      & 135.3                      & 197.6                      & 160.2                      & 123.2                      \\
+ SupCon                 & \textbf{154.5}             & \textbf{137.8}             & 261.1                      & 131.5                      & 190.1                      & 184.4                      & 127.8                      \\
+ DAI (ConPro)           & \textbf{154.5}             & \textbf{137.8}             & 233.1             & \textbf{118.0}             & \textbf{176.2}             & \textbf{155.8}             & \textbf{119.2}             \\ \hline
Joint (oracle)           & 42.9                       & 69.6                       & 77.1                       & 57.5                       & 65.1                       & 36.5                       & 106.7                  \\   
\Xhline{3\arrayrulewidth}
\end{tabular}}
\label{tab:main}
\end{table}

\begin{figure}[t]
  \includegraphics[width=0.85\linewidth]{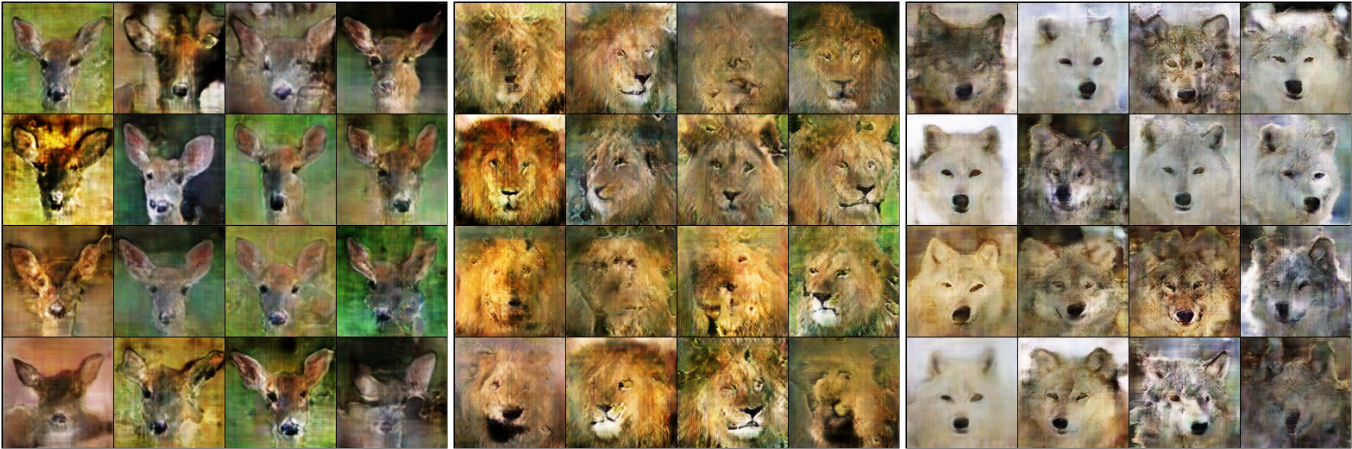}
   \vspace{-3mm}
  \centering
%   \captionsetup{width=0.8\linewidth}
  \caption{\small
  Uncurated (non-cherry-picked) samples from ConPro (100-shot).
  }
%   \vspace{-20pt}
\label{fig:qual}
% \vspace{-1.5em}
\end{figure}

% \vspace{-12pt}
\cref{tab:main} shows the quantitative results, where each component of ConPro contributes to the performance gain and sums up to provide significant advantage over the baseline. We note that our approach saves both parameters (\textbf{57.98M} vs 58.88M) and compute (30\% faster in wall-clock time). We present \textit{non-cherry-picked} samples from ConPro in \cref{fig:qual} for qualitative validations.

% \vspace{-1.5em}
\section{Conclusion}
\label{sec:con}

In this work, we tackled a relatively unexplored task of generative incremental few-shot learning. As it differs from simple IL and few-shot learning, we propose a novel GAN framework named ConPro that collaborately addresses the challenges.

\clearpage
\bibliographystyle{splncs04}
\bibliography{egbib}
\end{document}